\documentclass[a4paper,fleqn]{cas-dc}

\usepackage[numbers]{natbib}
\usepackage{adjustbox}
\usepackage{booktabs}
\usepackage{multirow}
\usepackage{svg}

\def\tsc#1{\csdef{#1}{\textsc{\lowercase{#1}}\xspace}}
\tsc{WGM}
\tsc{QE}
\tsc{EP}
\tsc{PMS}
\tsc{BEC}
\tsc{DE}

\begin{document}
\let\WriteBookmarks\relax
\def\floatpagepagefraction{1}
\def\textpagefraction{.001}
\shorttitle{Wall Shear Stress Estimation in Abdominal Aortic Aneurysms: Towards Generalisable Neural Surrogate Models}
\shortauthors{P. Rygiel et~al.}

\title [mode = title]{Wall Shear Stress Estimation in Abdominal Aortic Aneurysms: Towards Generalisable Neural Surrogate Models}                      



\author[1,2]{Patryk Rygiel}
[orcid=0009-0003-8539-5581]
\cormark[1]
\cortext[1]{Corresponding author}
\ead{p.t.rygiel@utwente.nl}
\credit{Conceptualization, Investigation, Methodology, Resources, Software, Visualization, Writing – original draft}

\author[3,4]{Julian Suk}
\credit{Conceptualization, Methodology, Writing – review and editing}

\author[1]{Christoph Brune}
\credit{Conceptualization, Supervision, Writing – review and editing}

\author[5,6,7]{Kak Khee Yeung}\credit{Conceptualization, Supervision, Data curation, Funding acquisition, Project administration, Writing – review and editing}

\author[1,2]{Jelmer M. Wolterink}
\credit{Conceptualization, Investigation, Methodology, Resources, Supervision, Funding acquisition, Writing – review and editing}

\affiliation[1]{
    organization={Department of Applied Mathematics, Technical Medical Centre, University of Twente},
    addressline={Drienerlolaan 5}, 
    city={Enschede},
    postcode={7522 NB},
    country={The Netherlands}
}

\affiliation[2]{
    organization={Cardiovascular Health Technology Centre, University of Twente},
    addressline={Drienerlolaan 5}, 
    city={Enschede},
    postcode={7522 NB},
    country={The Netherlands}
}

\affiliation[3]{
    organization={Department of Computer Science, Technical University of Munich},
    adressline={Boltzmannstr. 3},
    city={Garching},
    postcode={85748},
    country={Germany}
}

\affiliation[4]{
    organization={Munich Center for Machine Learning},
    adressline={Boltzmannstr. 3},
    city={Garching},
    postcode={85748},
    country={Germany}
}

\affiliation[5]{
    organization={Department of Surgery, Amsterdam UMC, Location University of Amsterdam},
    addressline={Meibergdreef 9}, 
    city={Amsterdam},
    country={The Netherlands}
}

\affiliation[6]{
    organization={Department of Surgery, Amsterdam UMC, Location Vrije Universiteit Amsterdam},
    addressline={De Boelelaan 5}, 
    city={Amsterdam},
    postcode={1083 HJ}, 
    country={The Netherlands}
}

\affiliation[7]{
    organization={Amsterdam Cardiovascular Sciences, Atherosclerosis \& Aortic diseases},
    city={Amsterdam},
    country={The Netherlands}
}



\begin{abstract}
Abdominal aortic aneurysms (AAAs) are pathologic dilatations of the abdominal aorta posing a high fatality risk upon rupture. 
Studying AAA progression and rupture risk often involves in-silico blood flow modelling with computational fluid dynamics (CFD) and extraction of hemodynamic factors like time-averaged wall shear stress (TAWSS) or oscillatory shear index (OSI).
However, CFD simulations are known to be computationally demanding.
Hence, in recent years, geometric deep learning methods, operating directly on 3D shapes, have been proposed as compelling surrogates, estimating hemodynamic parameters in just a few seconds.
A potential drawback of neural surrogates is that they are inherently data-driven and may suffer from common machine learning vulnerabilities such as poor generalisation to out-of-distribution samples. 
In this work, we propose a geometric deep learning approach to estimating hemodynamics in AAA patients, and study its generalisability to common factors of real-world variation. 

We propose an $E(3)$-equivariant deep learning model utilising novel robust geometrical descriptors and projective geometric algebra. 
Our model is trained to estimate transient WSS using a dataset of CT scans of $100$ AAA patients, from which lumen geometries are extracted and reference CFD simulations with varying boundary conditions are obtained. 
The model is validated using cross-validation and using an external test set consisting of $118$ AAA patients with reference CFD simulations.
We study the generalisation of this model to changing physiological conditions by varying boundary conditions, artery remodelling due to disease progression, varying artery tree topology, and changes in mesh resolution.

Results show that the model generalizes well within the distribution, as well as to the external test set. 
Moreover, the model can accurately estimate hemodynamics across geometry remodelling and changes in boundary conditions. 
Furthermore, we find that a trained model can be applied to different artery tree topologies, where new and unseen branches are added during inference. 
Finally, we find that the model is to a large extent agnostic to mesh resolution. 
These results show the accuracy and generalisation of the proposed model, and highlight its potential to contribute to hemodynamic parameter estimation in clinical practice. 
Our code is publicly available at https://github.com/PatRyg99/AAA-WSS-neural-surrogate.

\end{abstract}




\maketitle

\section{Introduction}
\label{sec:intro}
Cardiovascular diseases (CVDs) account for nearly $30$\% of all global deaths~\cite{lindstrom2022cvd,nabel2003cvd}.
Abdominal aortic aneurysms (AAAs) are a manifestation of CVD in which the abdominal aorta dilates more than $50$\% compared to its normal size~\cite{johnston1991aaa}.
Globally, AAA is estimated to affect $35$ million people aged $30$ to $79$~\cite{song2022prevelance}.
AAAs are typically asymptomatic; however, they come with a risk of rupture resulting in death in approximately $80$\% of cases~\cite{reimerink2013systematic}.
According to the European Society of Vascular Surgery (ESVS) guidelines, patients with an unruptured aneurysm whose diameter exceeds $5.5$ cm or $5.0$ cm, for men and women, respectively, are eligible for surgical intervention~\cite{wanhainen2024guidelines}.
These thresholds are chosen so that elective surgery is performed in those patients for whom the rupture risk exceeds the non-negligible risks of surgical complications~\cite{wanhainen2024guidelines}. 
However, the rupture risk is non-zero in AAAs with a diameter smaller than these thresholds, and only $5.3$\% of AAAs with a larger diameter rupture~\cite{darling1977autopsy,parkinson2015rupture}.
Hence, there is a need for improved stratification of AAA patients for elective repair.

Anatomical markers, such as the largest diameter, cannot fully describe AAA growth patterns and rupture risk.
As such, alternative, more reliable and quantifiable markers of AAA rupture have been explored~\cite{arnaoutakis2019abdominal,sweeting2012meta}.
In the search for such markers, biomechanics has shown promise as a way of obtaining functional markers of blood flow and tissue mechanics~\cite{arnaoutakis2019abdominal,vorp2007aaa}.
Wall shear stress (WSS), the tangential force exerted by blood flow on the inner vessel wall, plays a critical role in regulating endothelial cell (EC) function that makes up the inner lining of the vessel wall~\cite{reneman2006}.
Low or oscillatory WSS can promote EC inflammation that can lead to the build-up of atherosclerotic plaque~\cite{grego_endothelial_2025} or local weakening of the aortic wall and subsequent dilation or even rupture once the wall gets too weak~\cite{deroo2022}.
Consequently, combining such functional markers with anatomical ones could lead to improved growth and rupture risk assessment~\cite{alblas2025geometricdeeplearninglocal}.
Among biomechanical markers, peak wall stress derived from tissue mechanics~\cite{stevens2017biomechanical,gasser2014pws,martufi2016local}, and WSS-derived hemodynamic markers based on blood flow dynamics~\cite{stevens2017biomechanical,mutlu2023wss,boyd2016tawss,choke2005review,trenti2022wssmri,soudah2013aaacfd}, have been found to correlate with local aneurysm growth, site, and risk of rupture.
These markers can be determined via \textit{in-silico} modelling through computational methods~\cite{canchi2015review,manta2025review}.

In this study, we focus on hemodynamic parameters, which can be determined with computational fluid dynamics (CFD) methods.
CFD, a numerical approach to solving Navier-Stokes equations that govern fluid dynamics, is an established way of personalised blood flow modelling~\cite{taylor1998aaa}.
Although helpful in investigating spatiotemporal blood flow mechanics, CFD simulations come with a high computational burden in terms of both time and resources required, taking from minutes up to days, depending on the simulation complexity and the desired accuracy of the results. 
These high computational demands hamper the widespread use of CFD in the clinic, as well as the inclusion of hemodynamics in large-scale epidemiological studies.

To address the high computational costs of CFD, in recent years, deep learning (DL) methods have been proposed as surrogates for CFD solvers, achieving significant speed-ups while preserving the desired accuracy~\cite{taebi2022survey,arzani_machine_2022}.
Such surrogates are inherently data-driven and commonly follow a supervised learning regime.
One drawback of such data-driven methods is the requirement for representative training data, typically consisting of 3D geometries, boundary conditions, and corresponding CFD solutions.
As in any machine learning approach, such surrogate models can be susceptible to poor generalisation to shifts within and outside the distribution of the training data. 
Shifts within the data distribution may occur due to different physiological conditions, reflected in changing boundary conditions. 
Moreover, a patient's geometry can undergo remodelling over time due to disease progression.
Shifts between data distributions can occur due to changes to the artery tree topology or geometry. 
For example, artery trees encountered during inference might include branches not included during training, or meshes can have a different resolution. 
To be useful in the clinic or in large-scale studies, a model should be able to generalise to such changing conditions. 

Existing approaches to deep learning-based hemodynamics estimation project the artery tree's 3D geometry to 1D or 2D for use in fully-connected networks (FCNs)~\cite{liang2018pca,liang2020thoracic} or convolutional neural networks (CNNs)~\cite{ferdian2022wssnet,su2020cnn,yevtushenko2022centerline,faisal2025aaaunet}.
However, such approaches only allow limited generalisation to changes in topology and geometry, as they depend on hand-crafted parametrization schemes~\cite{liang2018pca,liang2020thoracic}, do not allow for variations in vascular topology~\cite{ferdian2022wssnet,su2020cnn,yevtushenko2022centerline}, or operate only on a partial geometry~\cite{faisal2025aaaunet}.
Alternatively, artery walls can be modelled as a continuous 2D manifold embedded in 3-dimensional Euclidean space and used as input to geometric deep learning (GDL) approaches. 
The 2D manifold describing an artery wall can be discretised as a 3D triangular mesh or point cloud.
GDL methods based on PointNet++~\cite{qi2017pointnet} and Transformer~\cite{vaswani2023} architectures have already been used successfully in surface-based (WSS~\cite{suk2024mesh,suk2024lab,brehmer2023geometric,lv2022tawss}, ECAP~\cite{morales2021ecap}) and volumetric-based (velocity~\cite{suk2024lab,Li2021aorta,suk2023velocity,suk2024physics,suk2024operator,zhang2023pinn}, pressure~\cite{Li2021aorta,suk2024operator,zhang2023pinn,nannini2025benchmark,rygiel2023}) estimation.
In addition, GDL methods can be further enhanced with geometrical priors by respecting Euclidean symmetries such as translation, rotation, and reflection in 3D space for improved data-efficiency compared to conventional methods~\cite{suk2023velocity}.

In this work, we hypothesise that a geometric deep learning approach allows training of flexible models that are, to a larger extent, robust to the aforementioned challenges. 
To verify this, we study the generalisability of neural surrogates for transient WSS estimation within AAA patients.
We propose to use a transformer-based architecture on a manifold discretised as a point cloud, operating either in linear or geometric algebra~\cite{suk2024lab}. 
We design robust input geometrical descriptors of the artery tree model, and perform extensive experiments to quantify the model's ability to adapt to shifts both within and between data distributions.  

\section{Materials \& Methods}
\begin{figure*}[t!]
    \centering
    \includegraphics[width=\textwidth]{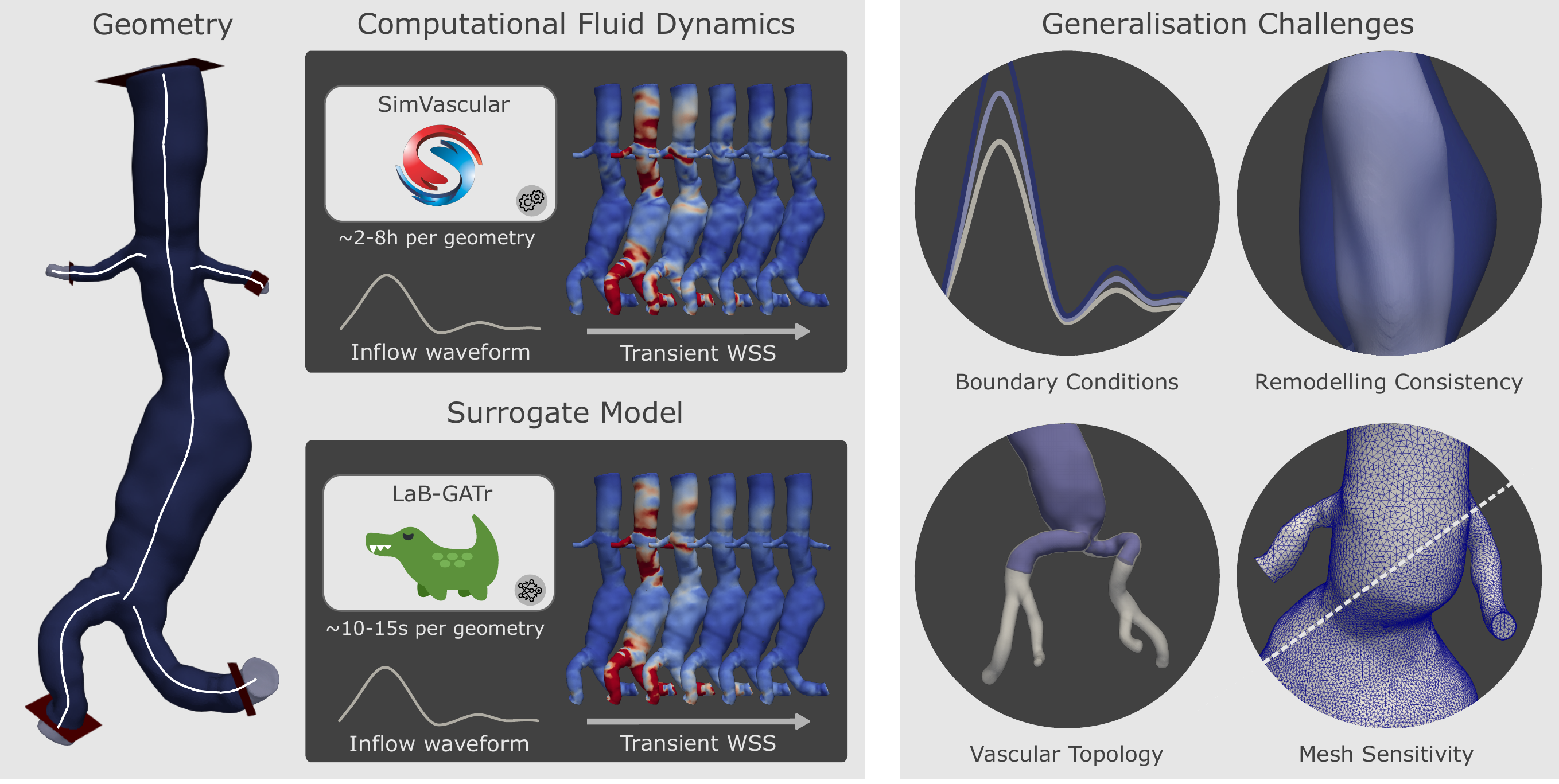}
    \caption{Workflow of hemodynamics estimation with CFD or deep-learning-based surrogate model. Both approaches take the CTA-obtained 3D geometry and inflow waveform as the input and yield transient WSS. Substituting the SimVascular CFD solver with a trained LaB-GATr model allows for $1500$x speedup in hemodynamics estimation. We study the generalisation capabilities of our model in four settings: changing physiological conditions due to varying boundary conditions, remodelling of artery geometry due to disease progression, changes in vascular topology during inference, and sentivity to mesh resolution during inference.}
    \label{fig:workflow}
\end{figure*}

We propose an automatic framework for deep learning-based transient WSS estimation in AAA patients. 
Fig.~\ref{fig:workflow} shows the workflow of the proposed approach.
A patient-specific geometry is extracted from a CT volume, and the inlet and outlet branches are defined.
This geometry can be used in either a conventional CFD pipeline or in the proposed surrogate model, using a patient inflow waveform.
In both cases, the obtained transient WSS field can be utilised to analyse patient hemodynamic quantities.
Our proposed approach is trained in a supervised manner with CFD solutions serving as a reference ground truth data.
After training, the model offers a $~1500\times$ speedup in hemodynamics estimation in comparison to traditional CFD.

To study the degree to which the proposed method generalises to shifts within and outside of the domain, we define \textit{generalisation challenges} that evaluate different aspects of method readiness to serve as a reliable neural CFD surrogate:

\paragraph{Boundary Conditions} - The hemodynamic behaviour depends on both vessel geometry and the prescribed boundary conditions.
The generalisable neural surrogate should offer decoupling of these input features and exhibit similar performance across different choices of boundary conditions.

\paragraph{Remodelling Consistency} - Vascular models undergo remodelling over time due to the disease progression.
The generalisable neural surrogate should be consistently accurate along such a longitudinal process to allow for reliable patient monitoring. 

\paragraph{Vascular Topology} - The vascular models may come with different branch configurations, either due to the patient-specific vasculature or differences in the acquisition region-of-interest.
The generalisable neural surrogate should offer zero-shot capability to estimate hemodynamics to a reasonable degree in such vasculatures.

\paragraph{Mesh Sensitivity} - Different discretisation schemes can represent the same vascular model.
The generalisable neural surrogate should be agnostic to the choice of the discretisation.

\subsection{Data}
We include two CT angiography (CTA) datasets referred to as \textit{AAA-100} and \textit{AAA-L}.
The AAA-100 dataset consists of pre-operative CTA volumes of $100$ AAA patients from Amsterdam UMC, Amsterdam, the Netherlands~\cite{rygiel2024aaa100}.
One CTA volume is included per patient.
The AAA-L dataset includes $29$ patients diagnosed with AAA at the Severance Hospital, Seoul, South Korea~\cite{kim2022baek1,jiang2020baek2}.
These patients had pre-operative CTA scans performed over up to $10$ years, resulting in $1 - 7$ scans per patient and a total of $118$ unique scans.

\subsubsection{Segmentation}
We automatically obtain a high-quality triangular surface mesh from CTA that describes the AAA lumen geometry. 
We employ the SIRE segmentation framework~\cite{rygiel2024global,alblas2025sire} that directly yields watertight surface meshes together with vessel centerlines, and allows for flexible branch inclusion.
We define a standardized AAA geometry that includes the abdominal aorta from the T12 vertebra until the iliac bifurcation.
Additionally, the geometry includes the first $5$ cm of both iliac arteries and the first $3$ cm of both renal arteries. 

For hemodynamic analysis, each geometry is volumetrically meshed and equipped with blood flow inlet and outlet planes. 
We automatically loft and cap the meshes based on the underlying centerlines and construct a tetrahedral mesh using a global edge length of $1.25$ [$mm$] with five boundary layers through the \texttt{TetGen} library~\cite{si2015tetgen}.
The constructed meshes consist of $1$ to $5.5$ million tetrahedral elements, depending on the aneurysm size.

\subsubsection{Computational Fluid Dynamics}
\label{sec:methods:cfd}
We train the deep learning-based CFD surrogate in a supervised manner, which requires CFD simulations as reference labels. 
We design an automatic pipeline that performs outlet boundary condition prescription, simulation, and post-processing.
We simulate blood flow through the 3D incompressible Navier-Stokes equations by employing the \texttt{svSolver} from SimVascular~\cite{updegrove2017simvascular}.
We assume dynamic viscosity $\mu=0.04$ [$\frac{g}{cm \cdot s}$], blood density $\rho=1.06$ [$\frac{g}{cm^3}$], rigid walls and no-slip boundary conditions.
As the inlet profile, we prescribe a template temporal abdominal aorta inflow waveform extracted from literature~\cite{taylor1998aaa} and a parabolic spatial profile.
For outlet boundary conditions, we prescribe a 3-element Windkessel model~\cite{taylor1998aaa,westerhof2009windkessel}.
We fix mean potential pressure $P_{\text{mean}}=125,000$ [$\frac{dyne}{cm^2}$], cardiac output $Q=80$ [$\frac{ml}{s}$], capacitance $C=0.001$ [$\frac{cm^5}{dyne}$] and distal pressure $P_{\text{d}}=0$.
The distal and proximal resistances are a fraction $0.91$ and $0.09$ of total resistance $R_{\text{total}} = \frac{P_{\text{mean}}}{Q}$, respectively.
To deal with multiple outlets, we split the resistances and capacitances according to the rules for a parallel circuit based on the areas of corresponding outlets~\cite{updegrove2017simvascular}.
The simulation is performed over four complete cardiac cycles, with each cycle lasting for $1$ [$s$], with a timestep of $0.002$ [$s$].
From the $500$ timesteps of the last cardiac cycle, we extract $21$ equitemporally distributed WSS fields across the cardiac cycle for model training and validation. 
For each geometry, a total of four distinct transient CFD simulations are performed by applying a scaled template inflow waveform~\cite{taylor1998aaa} (see Fig.~\ref{fig:embedding}), with peak systolic inflow $Q_{\text{max}}$ uniformly sampled between $60$ and $140$ [$\frac{ml}{s}$].
Additionally, for each geometry, a simulation with systolic inflow of $80$ [$\frac{ml}{s}$] is performed.
In total, we perform $1090$ simulations, each taking on average $5 \pm 3$ h on $32$ parallel processors on an AMD Rome CPU, accounting for approximately $227$ days of wall clock time. 

\subsection{Surrogate Model}
\label{sec:embedding}

We propose a geometric deep learning-based surrogate model for transient WSS estimation in AAA geometries. 
This requires a representation of the vascular geometry and a learnable way to operate on it.
Here, as a representation, we use the triangular surface meshes obtained automatically from CTA, with surface mesh $\mathcal{M} = (\mathcal{V}, \mathcal{F})$ consisting of vertices $\mathcal{V}$ and faces $\mathcal{F}$. 
The trained model should perform vector field regression in time to predict transient WSS. 
In our CFD formulation, WSS is independent of gravity and, thus, the orientation of the aorta in the ambient space.
We may assume that this model could be equivariant to translation, rotation, or reflection in 3D ambient space, i.e., equivariance with respect to the 3D Euclidean symmetry group $E(3)$.
Employing a model that respects those symmetries intrinsically is more data-efficient, by limiting the input space~\cite{suk2023velocity,brehmer2024doesequivariancematterscale}.

There are multiple ways to operate on 3D shapes and preserve these symmetries, e.g., Gauge-Equivariant Mesh CNN~\cite{haan2021gemcnn} or Steerable $E(3)$ Equivariant Graph Neural Network (SEGNN)~\cite{brandstetter2022segnn}.
Notably, the \textbf{g}eometric \textbf{a}lgebra \textbf{tr}ansformer (GATr)~\cite{brehmer2023geometric} and its variation, adapted to efficiently work on \textbf{la}rge-scale \textbf{b}iomedical meshes, LaB-GATr, currently hold state-of-the-art performance in WSS estimation in synthetic arterial models~\cite{suk2024lab}.
In this work, we use a LaB-GATr model, i.e., a transformer model that works on point clouds and exploits geometric algebra for $E(3)$ equivariance. 

\subsubsection{LaB-GATr}
The LaB-GATr architecture works on the point cloud representation of the geometry.
Hence, we strip the set of faces $\mathcal{F}$ to work only with the set of vertices $\mathcal{V}$ that, by convention, we refer to as a point cloud $\mathcal{P}$.
LaB-GATr is a wrapper around GATr, which consists of blocks that perform all operations within the geometric algebra $\mathbf{G}(3,0,1)$, naturally ensuring the preservation of all Euclidean symmetries.

Each point $p \in \mathcal{P}$ is described through $c$ geometric objects that are embedded through $16$-dimensional multi-vectors forming an embedding $X^{(0)} \in \mathbb{R}^{n  \ \times \ c \ \times \ 16}$ with $n = \vert \mathcal{P} \rvert$, that represents the full point cloud $\mathcal{P}$.
Given the embedding $X^{(l)}$, the transformer block is defined in the following manner:
 \begin{equation*}
    \begin{aligned}
        H^{(l)} &= \xi \left(\underset{h}{\lvert \rvert} \hspace{3pt} \text{Softmax} \left( \frac{q_h(X^{(l)}) k_h(X^{(l)})^T}{\sqrt{8c}}\right) v_h (X^{(l)})\right) \\[5pt]
        A^{(l)} &= X^{(l)} +  H^{(l)}, \quad X^{(l+1)} = A^{(l)} + \phi(A^{(l)})
    \end{aligned}
 \end{equation*}

where $q_h,k_h,v_h: \mathbb{R}^{n  \ \times \ d} \rightarrow \mathbb{R}^{n  \ \times \ d}$ are vertex-wise permutation-equivariant layers consisting of layer normalisation composed with learned linear maps.  
A multi-head self-attention $H^{(l)}$ is concatenated~($\lvert \rvert$) over $h$ heads, followed by a learned linear map $\xi$ and a geometric nonlinear layer $\phi$ with a residual connection.
All the operators in the GATr block work within geometric algebra $\mathbf{G}(3, 0, 1)$ resulting in equivariance to $E(3)$ symmetries of the input geometry represented by $X^{(l)}$: $\rho X^{(l)} \mapsto \rho X^{(l+1)}$.

To efficiently process large-scale meshes, LaB-GATr introduces a process of learned, geometric tokenisation through the cross-attention mechanism~\cite{suk2024labcross}.
Prior to passing $X^{(0)}$ to the GATr model, a coarse point cloud $\bar{\mathcal{P}} \subset \mathcal{P}$ is constructed by means of \textit{farthest-point sampling} with the Euclidean distance metric.
Subsequently, a coarse input embedding $\bar{X}^{(0)} \subset X^{(0)}$ corresponding to $\bar{\mathcal{P}}$ is retrieved.
The projection of a fine embedding onto the coarse one is performed via a multi-head cross-attention block that replaces the first multi-head GATr block:
 \begin{equation*}
    \begin{aligned}
        \bar{H}^{(0)} &= \xi \left(\underset{h}{\lvert \rvert} \hspace{3pt} \text{Softmax} \left( \frac{q_h(\bar{X}^{(0)}) k_h(X^{(0)})^T}{\sqrt{8c}}\right) v_h (X^{(0)})\right) \\[5pt]
        \bar{A}^{(0)} &= \bar{X}^{(0)} +  \bar{H}^{(0)}, \quad \bar{X}^{(l+1)} = \bar{A}^{(l)} + \phi(\bar{A}^{(l)})
    \end{aligned}
 \end{equation*}

Hence, wrapped GATr operates on a subsampled point cloud $\bar{\mathcal{P}}$, resulting in the final layer $L$ yielding a coarse embedding $\bar{X}^{(L)}$.
To recover the original point cloud resolution and the final fine prediction $Y$, the following learnt interpolation scheme is applied:
\begin{equation*}
    \begin{aligned}
        X^{(L)} \rvert_p &= \frac{\sum_{q \in \mathcal{N}(p)} \lambda_{p,q} \bar{X}^{(L)} \rvert_{q}}{\sum_{q \in \mathcal{N}(p)} \lambda_{p,q}},  \quad \lambda_{p, q} = \frac{1}{\lVert p - q \rVert^2_2 + \epsilon} \\[5pt]
        Y &= \psi(X^{L}, X^{0}) \in \mathbb{R}^{n  \ \times \ c \ \times 16}
    \end{aligned}
\end{equation*}

where $\rvert_p$ denotes the embedding of point $p$, $\mathcal{N}(p) \subset \bar{\mathcal{P}}$ is a set of $k$ nearest neighbors (in Euclidean distance sense) of point $p$ in $\bar{\mathcal{P}}$, $\epsilon$ is a small constant and $\psi$ is a geometric nonlinear layer.
Since we deal with surface meshes, we set $k = 3$~\cite{suk2024labcross}.

\subsubsection{Geometric Descriptors}
\begin{figure*}[!t]
    \centering
    \includegraphics[width=\textwidth]{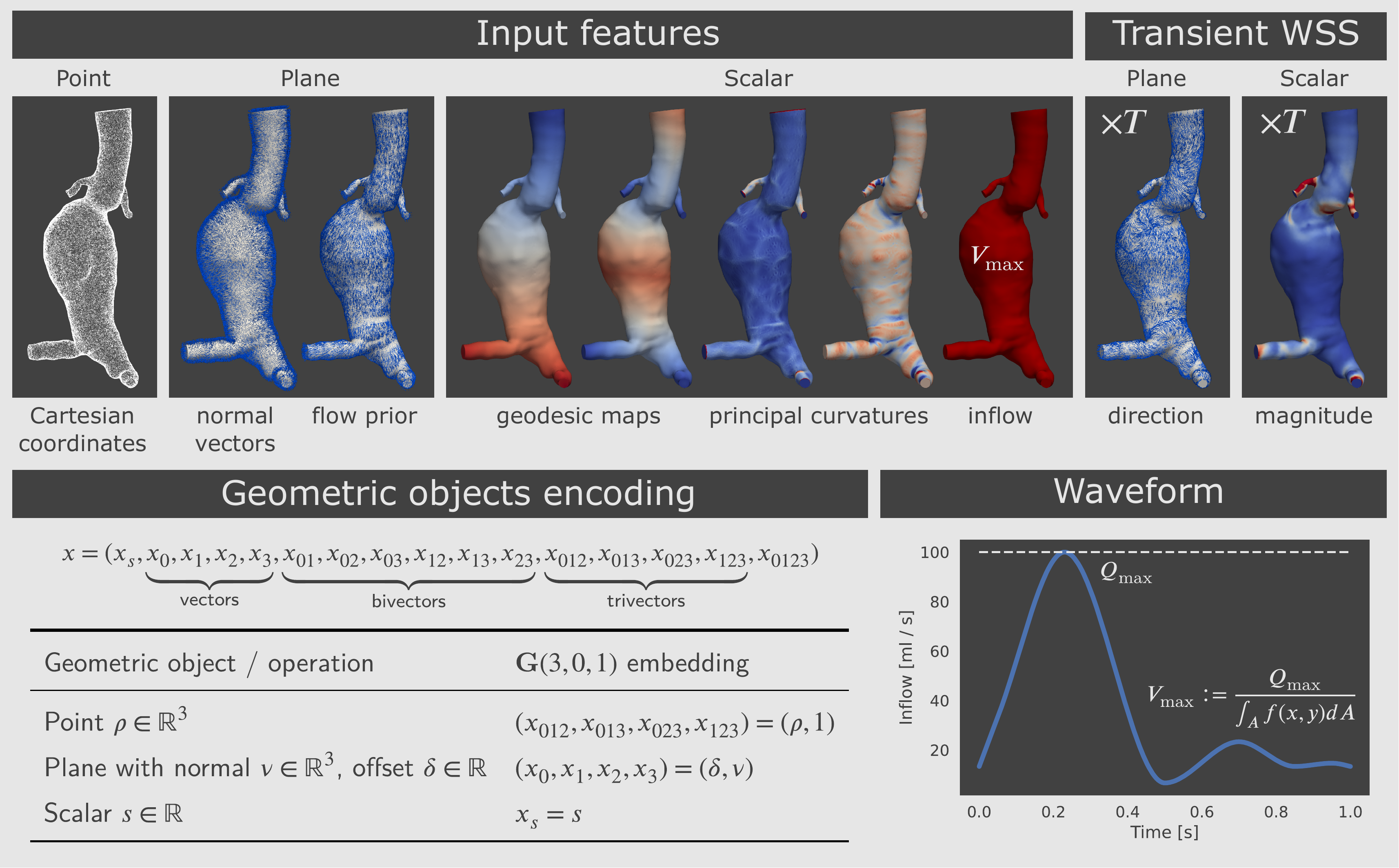}
    \caption{Input features of AAA geometry embedded through geometric objects of the geometric algebra $\mathbf{G}(3, 0, 1)$~\cite{suk2024lab,brehmer2023geometric}. Geometry encoded in such a manner serves as the input to the LaB-GATr model that respects all Euclidean symmetries, i.e., rotation, translation and reflection. The time-resolved inflow waveform is encoded by taking the maximum inflow $Q_{\text{max}}$ and computing maximum inlet velocity magnitude $V_{\text{max}}$ with inlet area $A$ and normalized spatial profile $f(x, y)$ (we assume a parabolic profile in all of our experiments).}
    \label{fig:embedding}
\end{figure*}
To aid LaB-GATr's expressiveness, it is crucial to design robust geometric descriptors and embed them correctly in $\mathbf{G}(3, 0, 1)$.
These descriptors should not break the model's intrinsic $E(3)$ equivariance.
We represent each point $p \in \mathcal{P}$ through the following point-wise geometrical descriptors (see Fig.~\ref{fig:embedding}):

\paragraph{Cartesian coordinates} - The Cartesian coordinates describe the shape and position of the geometry within the Euclidean space, and can be embedded in projective geometric algebra while maintaining all symmetries.

\paragraph{Normal vectors} - The normal vectors of the surface points define the orientation (inside-outside) of the geometry. 
The per-point normal vectors are obtained by averaging the normal vectors of adjacent faces.

\paragraph{Flow prior} - The direction of the WSS correlates with the direction of the flow within the vessel.
As such, we design WSS prior feature by approximating how the WSS would develop under the assumption of perfect laminar flow.
To compute it, we extract the direction of the inflow vector field, which is orthogonal to the inlet plane and points into the geometry.
Then, using the \textit{vector heat method}~\cite{sharp2019vectorheat}, we perform parallel transport of the inflow field across the whole surface.
The resulting feature is a tangential vector field, as is WSS itself, which approximately defines the local direction of the flow. \\

\paragraph{Geodesic maps} - A geodesic distance is defined as the distance along the manifold $\mathcal{M}$ between two points $x,y \in \mathcal{M}$.
We compute two such geodesic maps using the \textit{heat method}~\cite{crane2017heat}. 
The first one is the geodesic distance of each point to the vessel inlet.
The second one is the geodesic distance of each point to the closest vessel outlet.
Since the number of outlets can differ between geometries, we do not encode them separately but rather perform a minimum aggregation to form a single map representing all the outlets.
These maps provide an indication of the distance to the flow source and sink points within the domain.

\paragraph{Principal curvatures} - Curvature describes the local bending of the surface.
The two principal curvatures $\kappa_1, \kappa_2$ are the extreme values of all the normal curvatures and can be computed based on Gaussian curvature $\kappa_G$ and mean curvature $\kappa_H$ for each $p \in \mathcal{P}$ in the following manner~\cite{meyer2003curvature}:
\begin{equation*}
    \begin{array}{c}
        \kappa_1(p) := \kappa_H(p) + \Delta(p), \quad 
        \kappa_2(p) := \kappa_H(p) - \Delta(p) \\[5pt]
        \Delta(p) := \kappa^2_H(p) - \kappa_G(p)
    \end{array}
\end{equation*}

The time-dependent target that we're aiming to learn, i.e., transient WSS, is a marker defined on the surface. 
As such, we hypothesise that providing a measure of local surface behaviour might prove beneficial.
We compute both $\kappa_1$ and $\kappa_2$ through the $\kappa_G,\kappa_H$ operators in \texttt{PyVista}~\cite{sullivan2019pyvista}.

All proposed geometrical descriptors can be directly embedded using \textit{point}, \textit{plane} and \textit{scalar} geometric algebra $\mathbf{G}(3, 0, 1)$ objects (see~Fig.~\ref{fig:embedding}).
As such, all vector-valued features, i.e., normal vectors and flow prior, are embedded as \textit{plane} objects, and all the scalar-valued features, i.e., geodesic maps and principal curvatures, are embedded as \textit{scalar} objects, whereas the Cartesian coordinates are jointly embedded by means of a \textit{point} object.
All geometric descriptors are embedded separately into the multivectors, and concatenated together in the channel dimension to a single embedding $X \in \mathbb{R}^{n \ \times \ c \ \times \ 16}$, with $n$ being the number of geometry points, and $c$ the number of channels.
All the positions within a multivector not used for the object embedding are set to $0$ - see Fig.~\ref{fig:embedding} for the multivector structure.

\subsubsection{Inflow Conditioning}
To condition the network on patient-specific inflow parameters, we extend the set of input features to include the encoded inflow waveform.
Given an inflow waveform, we extract a peak inflow across time, denoted as $Q_{\text{max}}$ (see Fig.~\ref{fig:embedding}), and compute the magnitude of the maximal spatial velocity:
\begin{align*}
    V_{\text{max}} :=  \frac{Q_{\text{max}}}{\int_{A} f(x, y) dA}, \quad \underset{x \in X, y \in Y}{\text{max}} f(x, y) = 1
\end{align*}

where $A$ is the area of the inlet and $f(x, y)$ is a normalised 2D spatial inflow profile that we assume to be parabolic. 
This feature is constant for every geometry point and is encoded as a \textit{scalar} object in geometric algebra.

\subsubsection{Transient WSS}
LaB-GATr operates within the geometric algebra and yields as the output a set of multivectors from which the appropriate geometrical objects need to be extracted to obtain a transient WSS vector field $\tau \in \mathbb{R}^{T \ \times \ \vert \mathcal{P} \rvert \ \times \ 3}$, where $T$ number of simulation timepoints.
For each input point $p \in \mathcal{P}$ and timepoint $t \in \{1,..., T\}$, the model outputs a multivector $x^{\text{out}} \in \mathbb{R}^{16}$, from which we extract transient WSS predictions.

The transient WSS vector field $\tau$ can be decomposed into a scalar field $\bar{\tau} = \lVert \tau \rVert_2$ representing its magnitude and unit vector field $\hat{\tau} = \tau / \bar{\tau}$ representing its direction.
We employ this decomposition to extract appropriate geometric algebra objects from the output multivectors. 
For each output multivector $x^{\text{out}}$ we extract magnitude as a \textit{scalar} object $\bar{\tau}_{p,t} = x^{out}_s$ and unit vector as a normal of the \textit{plane} object $\hat{\tau}_{p,t} = (x^{out}_1, x^{out}_2, x^{out}_3)$.
The final predicted transient WSS field is constructed by composing both components.

\subsection{Model Training}
We train the LaB-GATr model using a loss function that decomposes the transient WSS vector field.
We employ L1 loss between the magnitudes and cosine similarity loss between the unit vector fields:
\begin{align*}
    L_{\text{magnitude}} &:= \underset{p \in \mathcal{P}, t = 1, ...,T}{\text{mean}} \ \lvert \lVert \bar{\tau}_{p,t}^{\text{true}} \rVert_2 - \lVert \bar{\tau}_{p,t}^{\text{pred}} \rVert_2 \rvert \\
    L_{\text{angle}} &:= 1 - \underset{p \in \mathcal{P}, t = 1, ...,T}{\text{mean}} \ \text{cos}\angle(\hat{\tau}_{p,t}^{\text{true}}, \hat{\tau}_{p,t}^{\text{pred}}) \\
    L_{\text{total}} &:= L_{\text{angle}} + \lambda L_{\text{magnitude}} 
\end{align*}

where $\mathcal{P}$ is a point could representing the geometry, $T$ is a number of predicted timepoints and \textit{pred}, \textit{true} refer to predicted and reference transient WSS, respectively.
We balance both terms by setting $\lambda = 0.1$.

\subsection{Quantitative Evaluation}
Both our model and the ground truth CFD directly yield transient WSS $\tau$ over $T$ timepoints.
We use it to derive two hemodynamic markers potentially associated with AAA growth and rupture site~\cite{stevens2017biomechanical,mutlu2023wss,boyd2016tawss,choke2005review}, i.e. time averaged wall shear stress (TAWSS) and oscillatory shear index (OSI); computed in the following manner for each $p \in \mathcal{P}$:
\begin{align}
    \label{eqn:tawss}
    \text{TAWSS} \ (p) &:= \frac{1}{T} \sum_{t=1}^T{\lVert \tau_{p,t} \rVert_2}  \\
    \label{eqn:osi}
    \text{OSI} \ (p) &:= \frac{1}{2} \left( 1 - \frac{\frac{1}{T} \lVert \sum_{t=1}^T{\tau_{p,t}} \rVert_2}{\text{TAWSS}(p)} \right)
\end{align}

Transient WSS and TAWSS are expressed in \textit{pascals} [PA] throughout this work, whereas OSI is \textit{unitless}, which we denote by [-]. 

To compare results to the CFD solutions, we compute four metrics: mean absolute error (MAE), normalised mean absolute error (NMAE), approximation disparity (Approx. disp.) and cosine similarity (Cos. similarity).
The cosine similarity is computed only for the transient WSS since TAWSS and OSI are scalar fields:
\begin{align*}
    \text{MAE} &:= \underset{p \in \mathcal{P}, t = 1, ...,T}{\text{mean}} \ \lVert \tau_{p,t}^{\text{true}} - \tau_{p,t}^{\text{pred}} \rVert_2& \\
    \text{NMAE} &:= \text{MAE} \ / \ \underset{p \in \mathcal{P}, t = 1, ...,T}{\text{max}} \ \lVert \tau_{p,t}^{\text{true}} \rVert_2& \\
    \text{Cos. similarity} &:= \underset{p \in \mathcal{P}, t = 1, ...,T}{\text{mean}} \ \text{cos}\angle(\tau_{p,t}^{\text{true}}, \tau_{p,t}^{\text{pred}})& \\
    \text{Approx. disp.} &:= \sqrt{\frac{\sum_{p \in \mathcal{P}, t = 1, ...,T}{\lVert \tau_{p,t}^{\text{true}} - \tau_{p,t}^{\text{pred}} \rVert^2_2}}{\sum_{p \in \mathcal{P}, t = 1, ...,T}{\lVert \tau_{p,t}^{\text{true}} \rVert^2_2}}}& \\
\end{align*}

\section{Experiments \& Results}
We train all models with $10$-fold cross-validation on AAA-100.
This set consists of $100$ shapes with $4$ distinct simulations each.
Hence, each model is trained on $90$ unique geometries ($360$ simulations in total).
Models trained in this cross-validation are combined into an ensemble, through a point-wise mean aggregation, and evaluated on an external test set AAA-L, which consists of $118$ unique geometries with $4$ distinct simulations each ($472$ simulations in total).

In the following sections, we present the experiments evaluating the proposed method's performance in the estimation of transient WSS and of derived hemodynamic markers TAWSS and OSI. 
Additionally, we study the method's generalisation capabilities.

Throughout this work, we use the same LaB-GATr architecture, consisting of $10$ blocks with $4$ heads each operating in the hidden dimension of size $8$. 
The tokenisation rate is set to $0.1$, i.e. $\lvert \bar{\mathcal{P}} \rvert = 0.1 \cdot \lvert \mathcal{P} \rvert$. 
Models are trained for $5000$ epochs with a batch size of $16$, using an \texttt{Adam} optimizer with a learning rate of $3e-4$, gradient clipping set to $1.0$, an exponential learning rate scheduler with decay parameter $\gamma = 0.9989$, and dropout ($p$=0.2) applied to geometric nonlinear layers $\phi,\psi$.
Training of a single model takes approximately $30$ hours on a single NVIDIA L40S/48G GPU. 

\subsection{Baseline Models}
\label{sec:baselines}
We compare the proposed approach to the following two baseline methods:

\paragraph{LaB-VaTr~\cite{suk2024operator}} follows the same transformer-based architecture design as Lab-GATr but uses linear instead of geometric algebra.
LaB-VaTr utilises the same set of features as LaB-GATr, but instead of encoding features as geometric objects, LaB-VaTr treats all of them as scalars and combines them into one single feature vector through concatenation.
The model follows the same training procedure as LaB-GATr but requires only $1000$ epochs until convergence, taking roughly $2.5$ hours on a single NVIDIA L40S/48G GPU.
In our experiments, LaB-VaTr consists of $12$ blocks with $4$ heads each operating in the hidden dimension of size $128$.
Additionally, we employ a dropout rate of $0.3$ that is applied to nonlinear layers $\phi,\psi$. The tokenisation rate is set to $0.1$, as it was done for LaB-GATr.

\paragraph{MultiViewUNet~\cite{faisal2025aaaunet}} is a 2D UNet architecture that operates on 2D snapshots of geometry taken from various angles. 
We chose this baseline method to represent a different school of hemodynamics estimation methods, which parametrise 3D shapes as 1D feature vectors or 2D images for use in an FCN or CNN. 
Most of these methods are not applicable in our experimental setting by being either limited to non-bifurcating vessels~\cite{ferdian2022wssnet,su2020cnn,yevtushenko2022centerline,faisal2025aaaunet} or vascular models with point-to-point correspondence~\cite{liang2018pca,liang2020thoracic}. 

MultiViewUNet takes a Gaussian curvature map as input and outputs a TAWSS and OSI map, both encoded as 2D RGB colour maps.
To condition the network on the inflow, we introduce a second input feature map $V_{\text{max}}$ which is constant for every point of the geometry. 
We follow the training procedure described in~\cite{faisal2025aaaunet} with a batch size of $32$ and $200$ epochs that takes $15$ hours on a single NVIDIA Quadro RTX 6000/24G GPU.

Note that MultiViewUNet does not provide a vector field but only a scalar field. Hence, we are limited only to the direct prediction of TAWSS and OSI. 
Retrieving 3D geometries from MultiViewUNet is non-trivial~\cite{faisal2025aaaunet}. 
Hence, as is done in~\cite{faisal2025aaaunet}, we compute NMAE on 2D snapshots (to which we refer to as snapshot normalised mean absolute error (Snap. NMAE)). 
For a fair comparison of methods, we render corresponding 2D snapshots for all the other methods that yield geometry predictions and evaluate them through Snap. NMAE as well.

\subsection{WSS Estimation}
\setlength{\tabcolsep}{0.5em}
{\renewcommand{\arraystretch}{1.2}%
    \begin{table*}[!t]
    \centering
    \caption[Short Heading]{We report \texttt{mean} $\pm$ \texttt{std} of hemodynamic parameter estimation. LaB-GATr and LaB-VaTr models directly predict transient WSS fields - TAWSS and OSI fields are computed based on these predictions according to equations~\ref{eqn:tawss},~\ref{eqn:osi}.}
    \begin{adjustbox}{width=\textwidth}
    \begin{tabular}{lllll|ll|ll}
    \hline
    \toprule         
         & & \multicolumn{3}{c}{\textbf{Transient WSS}} & \multicolumn{2}{c}{\textbf{TAWSS} } & \multicolumn{2}{c}{\textbf{OSI}}\\
         
        \multicolumn{1}{c}{\textbf{Dataset}} & \multicolumn{1}{c}{\textbf{Model}} & \multicolumn{1}{c}{\textbf{MAE} [$PA$] $\downarrow$} & \multicolumn{1}{c}{\textbf{Approx. disp} $\downarrow$} & \multicolumn{1}{c}{\textbf{Cos. similarity} $\uparrow$} & \multicolumn{1}{c}{\textbf{MAE} [$PA$] $\downarrow$} & \multicolumn{1}{c}{\textbf{Approx. disp} $\downarrow$} & \multicolumn{1}{c}{\textbf{MAE [-]} $\downarrow$} & \multicolumn{1}{c}{\textbf{Approx. disp} $\downarrow$} \\

        \midrule
        \multicolumn{1}{c}{\multirow{2}{*}{AAA-100$^\dagger$}} & 
        LaB-VaTr~\cite{suk2024operator} & $0.356 \pm 0.194$ & $0.770 \pm 0.144$ & $0.539 \pm 0.060$ & $0.187 \pm 0.102$ & $0.533 \pm 0.118$ & $0.084 \pm 0.010$ & $0.476 \pm 0.068$ \\
        & LaB-GATr~\cite{suk2024lab} & $\textbf{0.267} \pm 0.163$ & $\textbf{0.567} \pm 0.098$ & $\textbf{0.657} \pm 0.054$ & $\textbf{0.136} \pm 0.095$  & $\textbf{0.387} \pm 0.102$ & $\textbf{0.064} \pm 0.011$ & $\textbf{0.381} \pm 0.079$ \\ 
        
        \midrule
        \multicolumn{1}{c}{\multirow{2}{*}{AAA-L$^\ddagger$}} & 
        LaB-VaTr~\cite{suk2024operator} & $0.397 \pm 0.196$ & $0.674 \pm 0.109$ & $0.567 \pm 0.060$ & $0.229 \pm 0.144$ & $0.476 \pm 0.140$ & $0.074 \pm 0.009$ & $0.413 \pm 0.067$ \\
        & LaB-GATr~\cite{suk2024lab} & $\textbf{0.296} \pm 0.163$ & $\textbf{0.513} \pm 0.119$ & $\textbf{0.684} \pm 0.054$ & $\textbf{0.148} \pm 0.103$ & $\textbf{0.333} \pm 0.142$ & $\textbf{0.048} \pm 0.007$ & $\textbf{0.298} \pm 0.055$ \\ 

        
        \bottomrule
        \multicolumn{9}{l}{\small All models were trained on the AAA-100 dataset only} \\
        \multicolumn{9}{l}{\small $\dagger$ - cross-validation with $10$ folds} \\
        \multicolumn{9}{l}{\small $\ddagger$ - ensemble of all $10$ cross-validation folds} \\
        
    \end{tabular}
    
    \end{adjustbox}
    \label{tab:results-table}
    \end{table*}
}

\begin{figure*}[!ht]
    \centering
    \includegraphics[width=\textwidth]{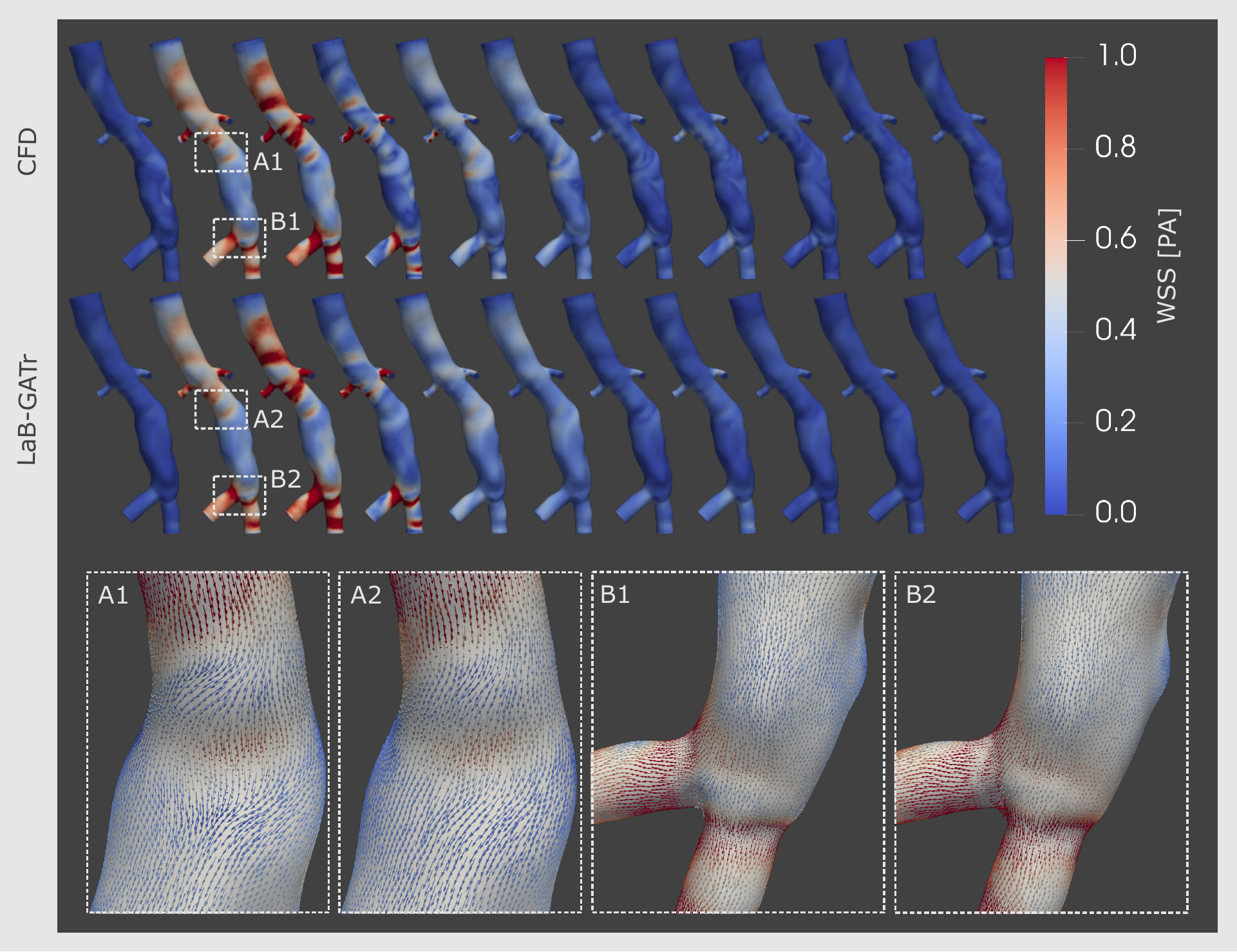}
    \caption{Qualitative comparison of CFD and LaB-GATr-obtained transient WSS field - for clarity we plot only every second timepoint. Based on transient WSS, TAWSS and OSI hemodynamic markers can be computed. Two regions are zoomed in to showcase the correspondence of predicted and ground truth WSS vector directions.}
    \label{fig:twss}
\end{figure*}
Table~\ref{tab:results-table} lists LaB-GATr and LaB-VaTr performance in transient WSS estimation. 
We found that both models are able to estimate transient WSS to a reasonable degree for the AAA-100 dataset. 
Moreover, we found that the LaB-GATr model outperforms the LaB-VaTr model, with MAE being lower by approximately $0.1$ [PA] when compared to the reference CFD. 
The Cos. Similarity is greater by roughly $0.1$, which corresponds to an improvement in angle agreement by $9^\circ$.
This shows the benefits of employing geometric algebra and its intrinsic E(3)-equivariance in architecture design.
Both models perform similarly on the external test set AAA-L as they do in the cross-validation experiment on AAA-100. 

Fig.~\ref{fig:twss} visualises qualitative results of LaB-GATr, showing the transient WSS magnitude across the cardiac cycle compared to CFD. 
We find that LaB-GATr achieved accurate spatial and temporal correspondence with respect to reference CFD.
The pulsatile pattern of WSS variation over the cardiac cycle was well captured, with high WSS values more pronounced at the time point of the maximum blood flow (see the applied template waveform in Fig.~\ref{fig:embedding}).
Insets A1, A2 and B1, B2 show the similarities between the CFD and deep learning-based WSS vector fields.
We can observe that the predicted direction of the vector field closely matches the reference one for both the straight vessel and the bifurcation.

Despite the model's overall performance, the CFD simulation exhibits some high-frequency patterns in transient WSS variation, which seem to be hard to capture by LaB-GATr, resulting in slight over-smoothing
This can also be observed in inset A in Fig.~\ref{fig:twss}, where the direction of the reference CFD WSS changes more rapidly than it does in the case of LaB-GATr.
These hemodynamic patterns arise due to the highly turbulent flow dynamics within the aneurysmal region.

\subsection{TAWSS \& OSI Estimation}
\begin{figure*}[!ht]
    \centering
    \includegraphics[width=\textwidth]{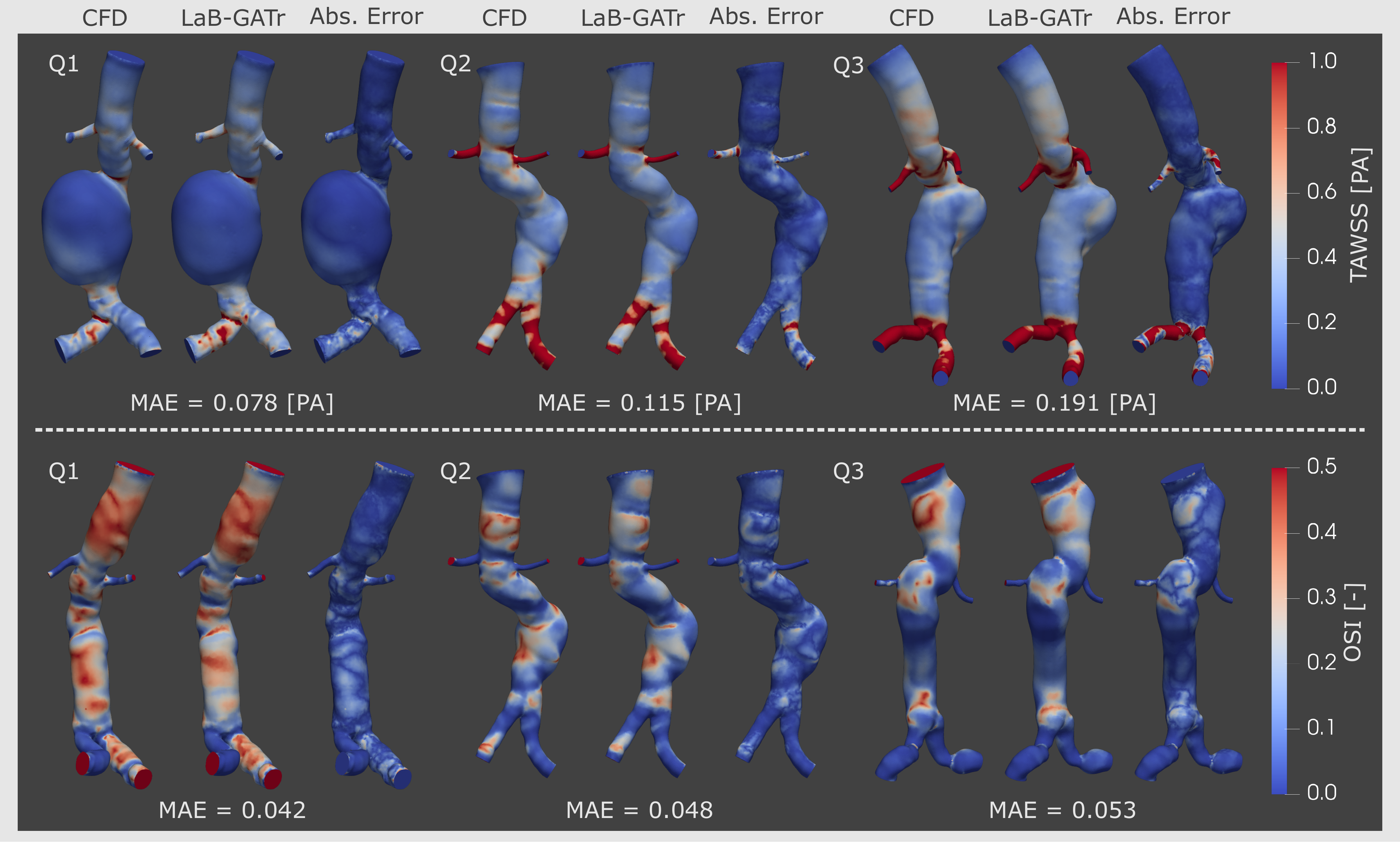}
    \caption{Qualitative comparison of CFD- and LaB-GATr-derived TAWSS and OSI fields. From the left, respectively, samples approximately at the first quartile (Q1), the second quartile (Q2 - median) and the third quartile (Q3) of the MAE distribution for TAWSS and OSI.} 
    \label{fig:tawss-osi}
\end{figure*}
\setlength{\tabcolsep}{0.5em}
{\renewcommand{\arraystretch}{1.2}%
    \begin{table}[!t]
    \centering
    \caption[Short Heading]{Comparison of deep learning methods: LaB-GATr, LaB-VaTr and MultiViewUNet. We utilise a Snap. NMAE metric to compare between the models and report its \texttt{mean} $\pm$ \texttt{std} for TAWSS and OSI.}
    \begin{adjustbox}{width=.5\textwidth}
    \begin{tabular}{lllll}
    \hline
    \toprule         
         & & \multicolumn{2}{c}{\textbf{Snap. NMAE} $\downarrow$} \\
         
        \multicolumn{1}{c}{\textbf{Dataset}} & \multicolumn{1}{c}{\textbf{Model}} & \multicolumn{1}{c}{\textbf{TAWSS}} & \multicolumn{1}{c}{\textbf{OSI}} \\

        \midrule
        \multicolumn{1}{c}{\multirow{3}{*}{AAA-100$^\dagger$}} & MultiViewUNet~\cite{faisal2025aaaunet} & $0.091 \pm 0.043$ & $0.209 \pm 0.031$ \\
        & LaB-VaTr~\cite{suk2024operator} & $0.079 \pm 0.037$ & $0.233 \pm 0.035$ \\
        & LaB-GATr~\cite{suk2024lab} & $\textbf{0.055} \pm 0.031$ & $\textbf{0.177} \pm 0.037$ \\ 
        
        \midrule
        \multicolumn{1}{c}{\multirow{3}{*}{AAA-L$^\ddagger$}} & MultiViewUNet~\cite{faisal2025aaaunet} & $0.108 \pm 0.051$ & $0.196 \pm 0.019$ \\
        & LaB-VaTr~\cite{suk2024operator} & $0.106 \pm 0.059$ & $0.213 \pm 0.037$ \\
        & LaB-GATr~\cite{suk2024lab} & $\textbf{0.066} \pm 0.041$ & $\textbf{0.135} \pm 0.027$ \\

        \bottomrule
        \multicolumn{4}{l}{\small All models were trained on the AAA-100 dataset only} \\
        \multicolumn{4}{l}{\small $\dagger$ - cross-validation with $10$ folds} \\
        \multicolumn{4}{l}{\small $\ddagger$ - ensemble of all $10$ cross-validation folds} \\
        
    \end{tabular}
    \end{adjustbox}
    \label{tab:snap-table}
    \end{table}
}

To further study the model's performance, we derive TAWSS and OSI from the transient WSS fields.
Table~\ref{tab:results-table} reports a quantitative evaluation of both. 
Note that TAWSS and OSI are scalar fields and we do not report cosine similarity.
Focusing on Approx. disp., we can observe that the performance gain of LaB-GATr over LaB-VaTr was higher in OSI estimation than in TAWSS estimation.
This observation supports the importance of equivariance in the model design, since OSI directly quantifies the vector field direction while TAWSS only quantifies its magnitude.

To further evaluate the performance of the LaB-GATr, Fig.~\ref{fig:tawss-osi} shows samples representing approximately the first, second, and third quartiles of the MAE error distribution for TAWSS and OSI, respectively.
Estimations of TAWSS are the most accurate across the samples in the aortic region, with the highest errors appearing in smaller branches - the iliac and renal arteries.
For the estimation of OSI, we can observe that the spatial distribution is quite well replicated.
However, there are intricate high-frequency patterns that are hard to capture.
Since OSI quantifies WSS direction variation, it inherently captures the turbulent flow patterns. 
The observation that turbulent patterns are hard to model echoes the over-smoothing problem present in transient WSS estimation.

Finally, to compare to non-geometric deep learning methods, for the estimation of TAWSS and OSI, we additionally employ a MultiViewUNet model~\cite{faisal2025aaaunet}.
Table~\ref{tab:snap-table} shows a comparison to the GDL models.
We can observe that MultiViewUNet achieved on-par performance with the LaB-VaTr model, being a little worse in TAWSS estimation and a bit better in OSI estimation. 
However, similarly to LaB-VaTr, the MultiViewUNet was outperformed by the LaB-GATr model.

\subsection{Feature Importance}
\setlength{\tabcolsep}{0.5em}
{\renewcommand{\arraystretch}{1.2}%
    \begin{table}[!t]
    \centering
    \caption[Short Heading]{We report \texttt{mean $\pm$ std} of transient WSS prediction for LaB-GATr trained with different input features. By "$+$" we denote cumulative addition of input features. We evaluate it on an AAA-L test dataset for a single model trained on AAA-100.}
    \begin{adjustbox}{width=.5\textwidth}
    \begin{tabular}{llll}
    \hline
    \toprule         
        \multicolumn{1}{c}{\textbf{Features}} & \multicolumn{1}{c}{\textbf{MAE} [$PA$] $\downarrow$} & \multicolumn{1}{c}{\textbf{Approx. disp} $\downarrow$} & \multicolumn{1}{c}{\textbf{Cos. similarity} $\uparrow$} \\

        \midrule
        Cartesian & $0.745 \pm 0.271$ & $1.173 \pm 0.246$ & $0.060 \pm 0.271$ \\
        + Normals & $0.392 \pm 0.187$ & $0.665 \pm 0.093$ & $0.614 \pm 0.053$ \\
        + Curvatures & $0.377 \pm 0.179$ & $0.637 \pm 0.082$ & $0.625 \pm 0.047$ \\ 
        + Geodesics & $0.366 \pm 0.181$ & $0.611 \pm 0.092$ & $0.626 \pm 0.052$ \\
        + Flow prior$^*$ & $\textbf{0.336} \pm 0.176$ & $\textbf{0.558} \pm 0.112$ & $\textbf{0.637} \pm 0.056$ \\

        \bottomrule
        \multicolumn{4}{l}{\small *Input features used in all the other experiments}
        
    \end{tabular}
    \end{adjustbox}
    \label{tab:ablation-table}
    \end{table}
}

We performed an ablation experiment in which we trained LaB-GATr models with different combinations of input features. 
Table~\ref{tab:ablation-table} lists results obtained on the external test set. 
We observe that representing the points by only their Cartesian coordinates is very limiting. 
Adding the normal vectors provides an almost two-fold improvement performance.
Including information about the geodesic distance to the inlet and outlets, together with principal curvatures, allows for additional improvement of $\sim 0.05$ for Approx. disp. and $\sim 0.1$ for MAE.
We observe the highest gain of $\sim 0.06$ Approx. disp. by adding the \textit{flow prior} feature.
This ablation proves that the design of informative geometry descriptors can further enhance the model performance even if the same amount of data is available.

\subsection{Generalisation Challenges}
We report results on the four \textit{generalisation challenges} that we identified for a neural surrogate model for CFD. 
The first two challenges, \textit{boundary conditions} and \textit{remodelling consistency}, evaluate how well the method performs under the shifts within the training data domain.
The last two challenges, \textit{vascular topology} and \textit{mesh sensitivity}, quantify the extrapolation capabilities by assessing how much the geometry can change beyond the training data domain without significant loss in the model's performance.

\subsubsection{Boundary Conditions}
\begin{figure}[h!]
    \centering
    \includegraphics[width=.5\textwidth]{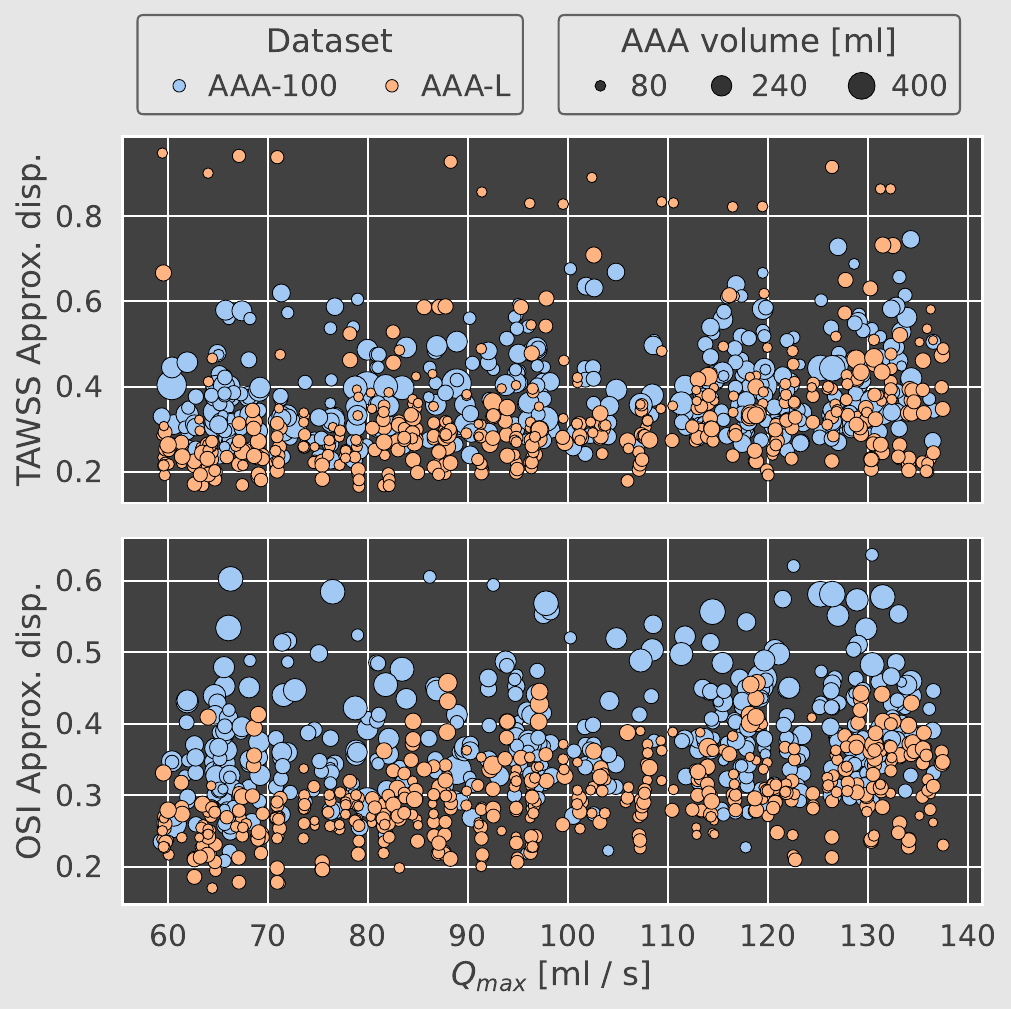}
    \caption{Distribution of Approx. Disp. for TAWSS and OSI across the samples with respect to the inflow rate and AAA volume.}
    \label{fig:aaa-scatter}
\end{figure}
In Fig.~\ref{fig:aaa-scatter}, we show TAWSS and OSI Approx. disp. distributions with respect to $Q_{\text{max}}$.
Both TAWSS and OSI estimation seem to be more difficult with the higher $Q_{\text{max}}$, with the OSI prediction error increasing more significantly.
Higher inflows lead to higher blood flow speed; hence, the Reynolds number is higher, meaning that blood flow patterns become more turbulent. 
Turbulence itself is quantified to a higher degree by the WSS direction variation rather than its magnitude.
Hence, the estimation of the OSI tends to be harder with higher blood inflow.

Additionally, in Fig.~\ref{fig:aaa-scatter} we mark cases by the size of AAA lumen volume.
The distribution of volume for TAWSS seems to be quite uniform with respect to the estimation error.
However, for OSI, there is a clear trend of larger AAA exhibiting a higher error rate.
Again, this observation can be tied to the difficulties of turbulence modelling.
The \textit{characteristic length} of AAA (i.e., its local radius), which is proportional to the Reynolds number, grows larger with increasing AAA volume, resulting in more turbulent flow patterns appearing in such geometries.

\subsubsection{Remodelling Consistency}
\label{sec:results:progression}
\begin{figure}[!t]
    \centering
    \includegraphics[width=.5\textwidth]{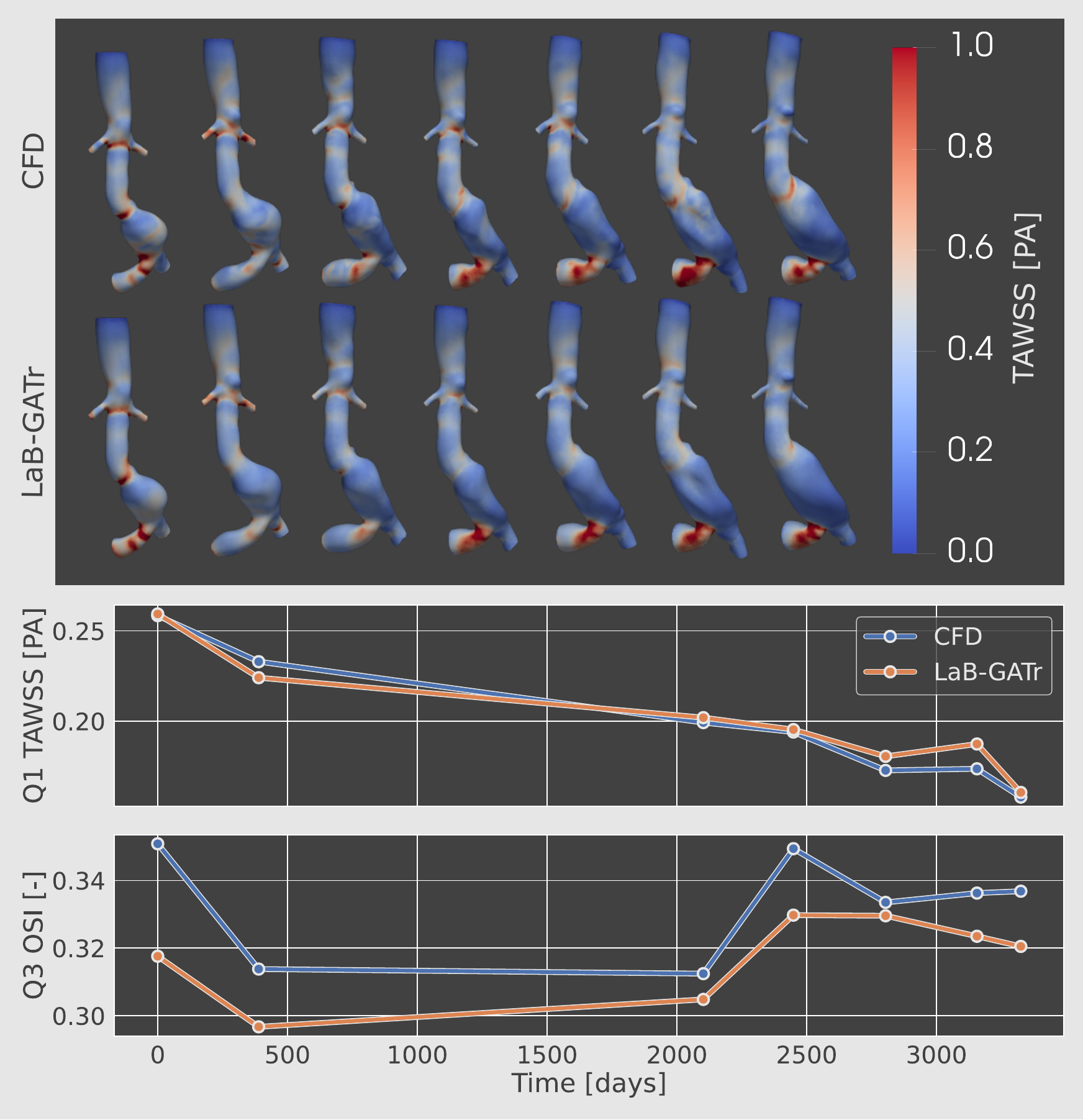}
    \caption{Consistency of Q1 TAWSS and Q3 OSI estimation over longitudinal patient imaging. We showcase a patient with $7$ longitudinal scans over the period of almost $10$ years.}
    \label{fig:longitudinal}
\end{figure}

Hemodynamic markers can be used to study the progression of AAA over time.
For example, low TAWSS and high OSI have been reported to correlate with local AAA growth~\cite{mutlu2023wss}.
Hence, a surrogate model should yield consistent estimates for the evolving vessel geometry over time, to allow for accurate longitudinal disease monitoring.
We consider such geometry shifts to be in-domain changes that require proper interpolation capabilities from the surrogate model.
We refer to this property of the model predictions being of consistent accuracy across single geometry changes as \textit{remodelling consistency}.

To study remodelling consistency, we utilise the external AAA-L test set that consists of $29$ patients with up to seven longitudinal scans per subject.
To study all longitudinal geometries under the same inflow conditions, we fix the maximum inflow rate $Q_{\text{max}}$ for all timepoints to be $80$~$\frac{ml}{s}$.
Since areas of low TAWSS and high OSI are usually of interest in AAA progression assessment~\cite{mutlu2023wss}, we calculate the first quartile (Q1) of TAWSS and the third quartile (Q3) of the OSI spatial distribution for each geometry.
We construct longitudinal trajectories per patient, representing their hemodynamic progression over time.
We compute the MAE between LaB-GATr and the reference CFD longitudinal trajectories for each patient separately. 
The reported mean and standard deviation across all the patients are $0.021~\pm~0.013$ [PA] and $0.020~\pm~0.010$ [-], for Q1 TAWSS and Q3 OSI, respectively.

Fig.~\ref{fig:longitudinal} shows TAWSS maps, as well as Q1 TAWSS and Q3 OSI values, over time in a patient who had seven CTA scans in ten years.
We can observe that both CFD and LaB-GATr can follow the pattern of TAWSS decreasing over time at the aneurysm surface, while the aneurysm expands.
The plotted Q1 TAWSS longitudinal trajectory conveys a similar pattern and depicts that the LaB-GATr estimation closely follows the reference CFD.
We can also observe that the LaB-GATr prediction is spatially more smoothed out than the CFD reference, missing out on the higher-frequency patterns.
The high-frequency patterns usually arise due to the turbulence, which is quantified to a higher degree by OSI than TAWSS.
Hence, the Q3 OSI longitudinal trajectory estimated by LaB-GATr is less accurate than that for TAWSS.
However, despite that, the estimated progression of hemodynamic markers is relatively consistent in time.
For the depicted patient, the OSI's maximal absolute error is $0.03$, which accounts for a maximal absolute normalised error of~$6\%$.

\subsubsection{Vascular Topology}
\setlength{\tabcolsep}{0.5em}
{\renewcommand{\arraystretch}{1.2}%
    \begin{table}[!ht]
    \centering
    \caption[Short Heading]{Qualitative comparison of zero-shot performance of LaB-GATr model on extended vascular geometries. We report \texttt{mean} $\pm$ \texttt{std} of transient WSS prediction. We denote abdominal aorta, common iliac and renal arteries as \textit{original} region; superior mesenteric artery, celiac trunk and both internal and external iliac arteries as \textit{added} region; and all the branches as the \textit{full} region.}
    \begin{adjustbox}{width=.5\textwidth}
    \begin{tabular}{lllll}
    \hline
    \toprule         
        \multicolumn{1}{c}{\textbf{Geometry}} & \multicolumn{1}{c}{\textbf{Eval. Region}} & \multicolumn{1}{c}{\textbf{MAE} [$PA$] $\downarrow$} & \multicolumn{1}{c}{\textbf{Approx. disp} $\downarrow$} & \multicolumn{1}{c}{\textbf{Cos. similarity} $\uparrow$} \\

        \midrule
        Original & Original & $0.179 \pm 0.066$ & $0.498 \pm 0.075$ & $0.697 \pm 0.036$ \\

        \midrule
        \multicolumn{1}{c}{\multirow{3}{*}{Extended}} & Original & $0.194^* \pm 0.064$ & $0.660^{**} \pm 0.078$ & $0.681^{\text{ns}} \pm 0.054$ \\ 
        & Added & $0.306 \pm 0.157$ & $0.582 \pm 0.091$ & $0.595 \pm 0.099$ \\
        & Full & $0.203 \pm 0.066$ & $0.650 \pm 0.071$ & $0.674 \pm 0.046$ \\

        \bottomrule
        \multicolumn{5}{l}{\small Wilcoxon test statistical significance with respect to original geometry:} \\
        \multicolumn{5}{l}{\small $\text{ns}$~- $0.05 < p \leq 1.0$} \\
        \multicolumn{5}{l}{\small $*$~~- $0.01 < p \leq 0.05$} \\
        \multicolumn{5}{l}{\small $**$~- $0.001 < p \leq 0.01$}
        
    \end{tabular}
    \end{adjustbox}
    \label{tab:topology-table}
    \end{table}
}
\begin{figure}[h!]
    \centering
    \includegraphics[width=.5\textwidth]{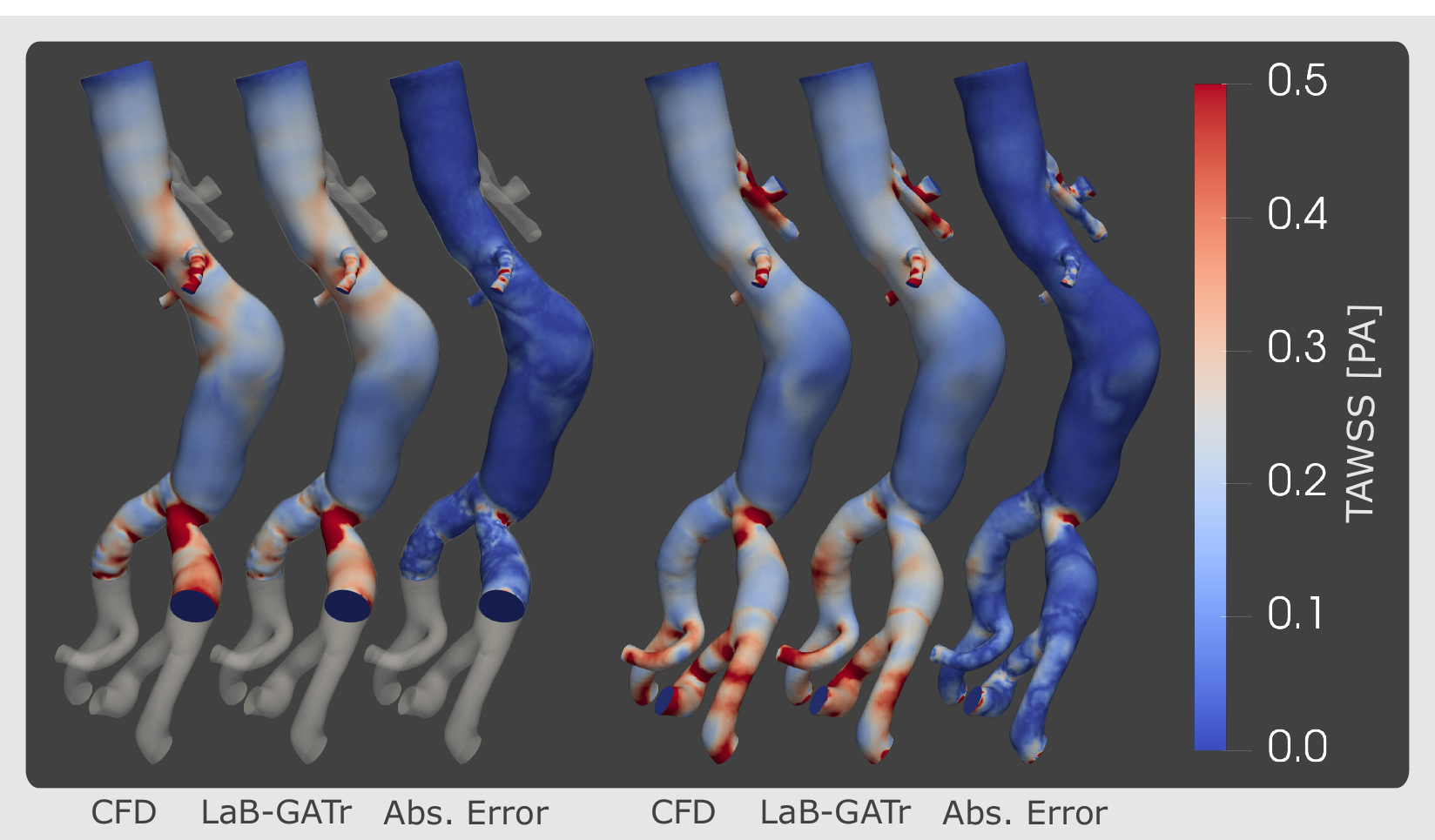}
    \caption{Qualitative comparison of TAWSS prediction between original and extended geometry with corresponding CFD-obtained ground truth.}
    \label{fig:aaa-topology}
\end{figure}
We study the extent to which a surrogate model can generalise to different \textit{vascular topologies}, i.e., different branch configurations.
Our surrogate model has been trained only on the AAA geometries encompassing the abdominal aorta, both common iliac and renal arteries.
Data-driven models tend to overfit to their training data and struggle to generalise when the input distribution changes — such as variations in vascular topology — leading to poor performance under domain shift.

Hence, to study the generalisation towards variation in \textit{vascular topology}, we construct a zero-shot evaluation scenario by extending a subset of AAA shapes to include different branch configurations than the shapes that we have trained the model on.
In a zero-shot scenario, the model is evaluated on out-of-domain samples without any retraining or changes to its architecture.
We utilise the SIRE framework~\cite{rygiel2024global,alblas2025sire} to extend $10$ shapes from the AAA-100 to additionally include the superior mesenteric artery, celiac trunk, and both internal and external iliac arteries beyond the iliac bifurcation.
We train the proposed model on $90$ AAA-100 geometries, leaving out $10$ original geometries that have been extended to avoid the potential risk of data leakage.
In these $10$ extended geometries, the CFD simulations have been performed in the same manner as described in Section~\ref{sec:methods:cfd}.
We fix peak systolic inflow $Q_{\text{max}}$ for all the original and extended geometries to be $80$~$\frac{ml}{s}$. 

Table~\ref{tab:topology-table} lists a comparison of model performance between the CFD and LaB-GATr in these geometries.
We report metric scores for different evaluation regions, namely, the branch configuration on which the model was originally trained (\textit{original}), the branches that were added to extend the geometry (\textit{added}), and the \textit{full} extended geometry.
Looking at the original evaluation region, we can observe that MAE and Approx. disp. deteriorate significantly ($p < 0.05$ in the Wilcoxon test) between the original and the extended model.
However, the differences are not large, with MAE dropping only by $0.015$ [PA] and Approx. disp. by $0.162$.
The only metric in which there is no significant deterioration is Cos. Similarity.
This means that the evaluation of WSS direction is more reliable than its magnitude once the model is extended.
Zero-shot performance on newly added branches exhibits much higher MAE; however, one needs to consider that the WSS magnitude is, on average, much higher in these smaller branches than in the original.

Fig.~\ref{fig:aaa-topology} shows a qualitative comparison of TAWSS between CFD and LaB-GATr for both original and extended geometries.
In line with the quantitative analysis, we can see that the zero-shot performance on extended geometry corresponds closely to the reference CFD in the original area.
While predictions on new branches are more prone to being erroneous, as depicted by the absolute error map.
Interestingly, the CFD-obtained TAWSS distribution varies between the two geometries in the original region.
This is because new outlets have been added to the simulation, and thus, the outlet RCR boundary conditions have changed, leading to a slightly different solution.
We observe similar behaviour for LaB-GATr, which, in a zero-shot manner, can adapt to the change of boundary conditions by the inclusion of additional branches.
This zero-shot capability is desired since it reveals that the model has grasped the underlying nature of the hemodynamic behaviour with respect to the geometry and can extrapolate beyond the training data distribution.

\subsubsection{Mesh Sensitivity}
\begin{figure*}[!t]
    \centering
    \includegraphics[width=\textwidth]{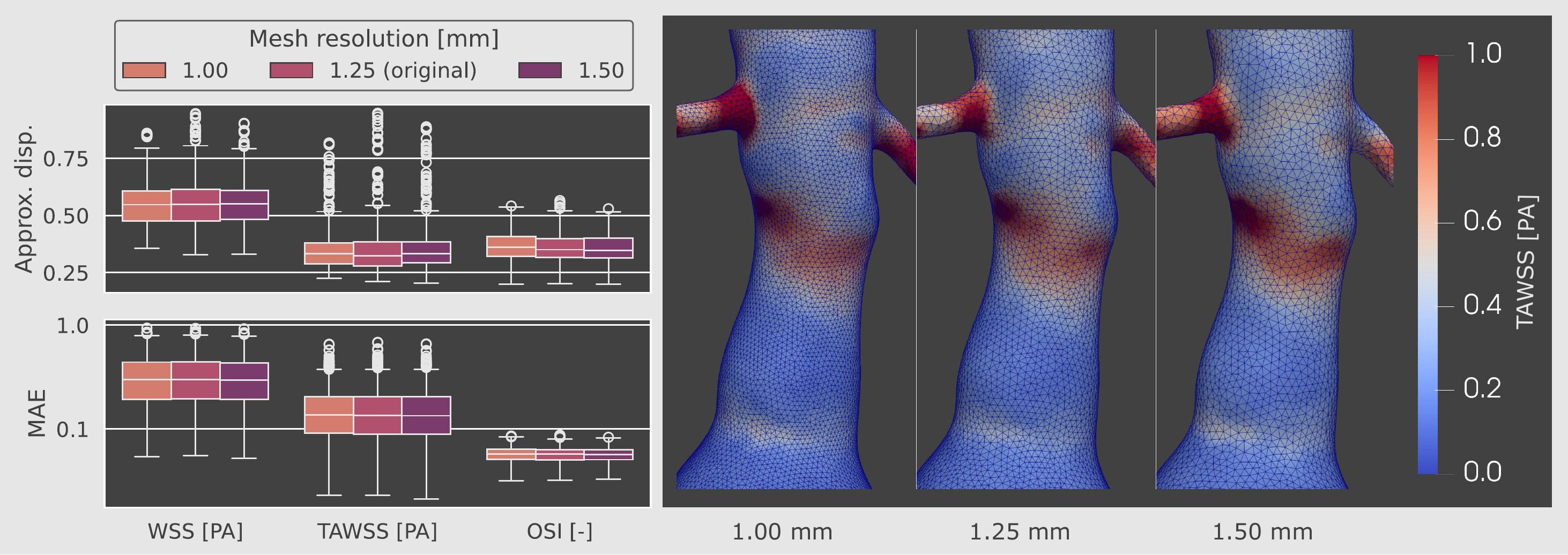}
    \caption{Consistency of LaB-GATr predictions across different mesh resolutions. The model was trained on meshes with a resolution (global edge length) of $1.25$ mm - denoted with \textit{original}. Visual comparison shows a predicted TAWSS field with the underlying surface discretisations for different resolutions.}
    \label{fig:remeshing}
\end{figure*}

We consider vascular geometry as a continuous 2D manifold embedded in a 3D space.
However, for the sake of numerical solving or AI-based analysis, it needs to be discretised to a finite set, such as a mesh or a point cloud.
The choice of discretisation should not inhibit performance, as it is merely an artefact of transitioning from a continuous to a finite representation.
Moreover, in real-world scenarios, such different discretisations may arise due to a number of pre-processing factors

To study to what degree the proposed method is agnostic to the discretisation of the vessel wall surface, we perform a \textit{mesh sensitivity} analysis.
The original models were trained on a mesh with a global edge size of $1.25$ mm - referred to as \textit{original}.
We remesh the AAA-L test set to $1$ mm and $1.5 $mm resolutions, and evaluate the model performance on these more- and less-detailed meshes.

In Fig.~\ref{fig:remeshing}, we present box plots representing error distributions for WSS, TAWSS and OSI between different meshing resolutions.
We observe that the errors follow the same distribution and the differences are not statistically significant with respect to the original resolutions ($p>0.05$ in a Wilcoxon test).
Next to the box plots, the visual comparison between different resolutions is shown.
While we can spot some differences, especially around renal artery inlets, they are natural artefacts of detail over-smoothing caused by the sparser surface representation.

\section{Discussion}
We have proposed a geometric deep learning-based model for transient WSS estimation in AAA patients, and studied its generalisation to changes in geometry, topology, and physiology. 
In a series of experiments, we have shown that the neural surrogate model, based on the LaB-GATr architecture, can be successfully applied to unseen AAA anatomies. 
The model appears robust to within-domain shifts by allowing accurate querying of WSS for patient-specific physiological conditions realised through the inflow waveform, and offering a similar quality of patient hemodynamic longitudinal tracking as the reference CFD. 
Additionally, the model can be successfully applied to out-of-domain geometries by extrapolating in a zero-shot manner to geometries of unseen topology and different meshing resolutions.

Our study shows how deep learning methods that operate on the artery wall manifold and exploit symmetries can effectively learn from relatively small data sets and generalise well to new conditions. 
In contrast, many previously proposed deep learning models by design do not offer this flexibility. 
They are limited by specific geometry reparameterisation, requiring either point-to-point correspondence between geometries~\cite{liang2018pca,liang2020thoracic} or fixed vascular topology~\cite{ferdian2022wssnet,su2020cnn,yevtushenko2022centerline}. 
We also found that while MultiViewUNet~\cite{faisal2025aaaunet}, a projection-based CNN used for TAWSS estimation in AAA, is 
capable of estimating TAWSS and OSI on projected views, its performance is substantially worse than that of LaB-GATr. 



The utilised LaB-GATr architecture is $E(3)$-equivariant by operating within projective geometric algebra. 
Preservation of symmetries is beneficial in scenarios where the training data is scarce~\cite{brehmer2024doesequivariancematterscale}. Transient WSS can be modelled as an $E(3)$-equivariant vector-valued field since it is inherently intrinsic to the geometry due to external forces such as gravity being ignored in reference CFD. 
Hence, the model learns to correlate the transient WSS with changes in local geometry and not rely on spurious features such as absolute location in ambient space.  
Almost as important as the model are the features that we provide it. 
We found the flow prior feature, which approximately corresponds to blood flow direction under the perfect laminar flow assumption, to be an adequate geometrical descriptor when learning WSS. 
Moreover, we found that by making geodesic maps agnostic to the number of artery outlets, we could achieve generalisation to different artery tree topologies. 
Finally, we found that the way in which features are embedded is important. 
The LaB-GATr model exploits scalar and vector-valued geometrical descriptors that are expressively embedded through the projective geometry algebra objects, rather than being concatenated into a 1D vector, as is required in the case of linear algebra. 
The performance gain we found for LaB-GATr over other models echoes that found in a recent benchmark on fractional flow reserve estimation in coronary arteries~\cite{nannini2025benchmark}.

We have utilised the predicted transient WSS to derive two hemodynamic markers, TAWSS and OSI.
These markers serve a clinical purpose by being reported to correlate with AAA local growth and site of rupture~\cite{mutlu2023wss}, as well as an evaluation purpose by quantifying different aspects of transient WSS.
TAWSS is a descriptor of transient WSS magnitude only, while OSI quantifies mostly the variation in directionality over the cardiac cycle.
We observed that both marker fields were usually more smoothed out, missing high-frequency patterns, when compared to the reference CFD, with the TAWSS being estimated with higher quality than the OSI.
These shortcomings are the result of imprecise turbulence estimation by the surrogate model.
The turbulent patterns of WSS are inherently driven by the velocity patterns forming within the dilated AAA geometry.
We hypothesize that, due to the utilised surface-only approach, the training supervision given by the surface-bound WSS reference is not exhaustive enough.
A natural extension of the proposed approach would be the modelling of the full velocity and pressure field.
However, such an approach would require switching from surface- to volumetric-mesh representation, resulting in much higher computational complexity, from tens of thousands to millions of points.
Scaling up neural surrogates to millions of points remains a challenge that has started to be addressed by recent works employing conditional neural fields~\cite{alkin2024upt,bleeker_neuralcfd_2025}.

In recent years, there has been growing interest in the use of data-driven methods to (partially) replace time-consuming CFD simulations~\cite{arzani_machine_2022,taebi2022survey}. However, in order for these methods to be easily deployable and adaptable in the clinic, they should generalise well under distribution shifts within and outside of the domain they were trained on.
We performed a thorough empirical evaluation of the model’s robustness in four settings that can occur in real-world implementations, which we have referred to as \textit{generalisation challenges}.

Hemodynamic markers depend both on vessel geometry and the prescribed boundary conditions.
Hence, we studied in-domain generalizability through changing \textit{boundary conditions} and \textit{remodelling consistency}.
We observed that the model can be successfully conditioned on the inflow volume to be adapted to patient-specific physiological conditions.
We also observed that performance slightly deteriorates, particularly for OSI estimation, as inflow volume increases.
Regarding the geometry, we have observed that the estimation of OSI tends to be also more difficult in larger aneurysms, while TAWSS seems not to be that sensitive to this feature. 
Both of these observations regarding the OSI estimation can be tied to an increase in Reynolds number.
Hence, a shift of flow from a more laminar to a more turbulent regime.
The latter one, due to its chaotic nature, is harder to estimate with the same accuracy as the former one.
As such, neural surrogates may require a higher volume of training data for turbulent cases to exhibit similar performance across the full domain.

In addition to the changes in AAA volume, we also studied single-patient geometry remodelling over time. The local progression of AAA has been tied to trends such as increasing OSI and decreasing TAWSS~\cite{mutlu2023wss}.
We showed that the neural surrogates exhibit similar tracking capabilities to these markers as reference CFD.
As such, they offer a clinical value as a longitudinal patient monitoring tool.

We studied the model's zero-shot capabilities under changes in \textit{vascular topology}.
We observed that the developed surrogate is capable of estimating WSS in geometries of unseen branch configurations, with new branches proving to be more difficult than the original region.
It is worth noting that due to the addition of new branches, not only the geometry but boundary conditions as well have not been seen before by the surrogate.
Due to the change in boundary conditions, the reference hemodynamics within the original region between the original and extended geometry have changed.
Our surrogate managed to fairly accurately estimate this boundary condition shift with respect to reference CFD.
This provides insight into how the model effectively grasped the underlying laws of hemodynamic behaviour.
To further enhance the surrogate performance for new geometries and boundary conditions, while keeping with the data-efficiency paradigm, methods utilising transfer learning could be employed~\cite{bleeker_neuralcfd_2025}.

Another form of out-of-distribution shift is the change in a geometry discretisation, which we study through \textit{mesh sensitivity}.
In principle, the model's performance should not depend on the choice of discretisation, since all the discretisations represent the same geometry.
We perform a zero-shot evaluation on geometries of higher and lower resolution and show that the model performance is indeed agnostic to the underlying mesh resolution. 
This is a desired but by no means trivial quality, since models operating on the mesh connectivity tend to overfit to its structure, rendering them unreliable once the same vasculature is represented with a different discretisation~\cite{suk2024mesh}.
By disregarding the connectivity, the proposed model learns the underlying continuous 2D manifold through the point cloud representation, and is able to accurately interpolate points in-between, once geometry is represented by more or less points.

We have studied the generalisation challenges in isolation; however, it is essential to note that in practice, they are likely to occur simultaneously. 
For example, in our \textit{remodelling consistency} experiments, we assumed a constant inflow across all the timepoints.
However, in practice, the patient's physiological conditions may change over time.
The errors regarding different generalisation capabilities may compound, leading to larger performance deterioration once both patient geometry and boundary conditions move further from the training data domain.
The construction of a representative test set that allows for the study of these errors separately and in a compound manner is mandatory.
Moreover, we find it crucial to report errors for different aspects of neural surrogate separately to indicate their certainty to various domain shifts, and hence, provide the bounds of their clinical reliability.

By modelling full transient WSS, many other markers, such as relative residence time (RRT) or endothelial cell activation potential (ECAP), could be derived analogously~\cite{mutlu2023wss}.
Moreover, a neural surrogate can be developed for any other biomechanical factors, such as peak wall stress~\cite{gasser2014pws}, by retraining on the data derived from that biomechanical model.
Studying other biomechanical markers might require the inclusion of other anatomical structures, such as intraluminal thrombus (ILT), which could be obtained automatically~\cite{alblas_going_2022}.

Our study has limitations. 
First, ideally, hemodynamic biomarkers are computed using personalised boundary conditions. 
However, in our study, we did not have access to any patient-specific flow measurements to prescribe specific boundary conditions.
Instead, we derived the template inflow waveform from the literature. 
However, once the patient-specific training data is available, the model could be retrained or fine-tuned to reflect the tuned biomechanical model.
Second, while we used a relatively large set of $100$ AAA patients to train and validate our model, and exploited data-efficient geometric deep learning techniques, it is likely that larger training sets could lead to improved results. 
To address this problem, in addition to symmetry-preservation~\cite{suk2023velocity}, other techniques such as transfer learning~\cite{bleeker_neuralcfd_2025} or active learning~\cite{rygiel2025activelearning} could be further explored.

In conclusion, we have proposed a neural surrogate for transient WSS estimation in AAA utilising a geometric algebra transformer and geometric descriptors that exhibit a variety of generalisation capabilities. 
This model has the potential to contribute to personalised hemodynamic parameter estimation.

\section{Acknowledgements}
We would like to thank dr. Kartik Jain for the discussions on approaching building the CFD pipeline for patient geometries, and authors of~\cite{kim2022baek1,jiang2020baek2} for sharing the CTA data of AAA patients that served in the construction of the AAA-L dataset.
This work made use of the Dutch national e-infrastructure, in particular the Dutch supercomputer Snellius, with the support of the SURF Small Compute Applications grant no. EINF-10199.
This project has received funding from the European Union's Horizon Europe research and innovation programme under grant agreement No 101080947 (VASCUL-AID).



\printcredits

\bibliographystyle{cas-model2-names}
\bibliography{bibliography}

\end{document}